\documentclass[11pt]{article}

\usepackage[preprint]{acl}
\usepackage{lmodern}
\usepackage{fix-cm}

\usepackage{times}
\usepackage{latexsym}
\usepackage{enumitem}
\usepackage[table]{xcolor}
\usepackage{booktabs}
\usepackage{bbm}
\usepackage{dsfont}
\usepackage{multicol}
\usepackage{multirow}
\usepackage{algorithm}
\usepackage{algorithmic}
\usepackage{amsmath}
\usepackage{amsfonts}
\usepackage{graphicx}

\usepackage{tcolorbox}
\tcbuselibrary{skins, breakable}

\definecolor{blue}{RGB}{138,110,172}
\definecolor{bluelight}{RGB}{241,239,246}
\definecolor{blueband}{RGB}{223,234,246}

\newcommand{\camphead}[1]{\textcolor{white}{\textbf{#1}}}
\newcommand{\campmetric}[1]{\textbf{#1}}

\tcbset{
  promptbox/.style={
    enhanced,
    breakable,
    colback=bluelight,
    colframe=blue,
    fonttitle=\bfseries\small,
    coltitle=white,
    colbacktitle=blue,
    boxrule=0.6pt,
    arc=2pt,
    left=6pt, right=6pt, top=10pt, bottom=6pt,
    before upper={\small},
    attach boxed title to top left={xshift=4pt, yshift=-\tcboxedtitleheight/2},
    boxed title style={size=small, colback=blue, colframe=blue, arc=1.5pt},
  },
  casebox/.style={
    enhanced,
    enforce breakable,
    colback=white,
    colframe=blue,
    fonttitle=\bfseries\small,
    coltitle=white,
    colbacktitle=blue,
    boxrule=0.6pt,
    arc=2pt,
    left=6pt, right=6pt, top=6pt, bottom=6pt,
    before upper={\small},
  },
  votebox/.style={
    enhanced,
    breakable,
    colback=bluelight,
    colframe=blue!50!white,
    boxrule=0.4pt,
    arc=1.5pt,
    left=5pt, right=5pt, top=3pt, bottom=3pt,
    before upper={\small},
  },
  conflictbox/.style={
    enhanced,
    breakable,
    colback=red!4!white,
    colframe=red!40!blue,
    boxrule=0.4pt,
    arc=1.5pt,
    left=5pt, right=5pt, top=3pt, bottom=3pt,
    before upper={\small},
  },
  distractorbox/.style={
    enhanced,
    breakable,
    colback=white,
    colframe=black!30,
    boxrule=0.4pt,
    arc=1.5pt,
    left=5pt, right=5pt, top=3pt, bottom=3pt,
    before upper={\small},
  },
}

\usepackage[T1]{fontenc}

\usepackage[utf8]{inputenc}

\usepackage{microtype}

\usepackage{inconsolata}

\usepackage{graphicx}

\title{One Panel Does Not Fit All: Case-Adaptive Multi-Agent Deliberation for Clinical Prediction}

\author{
  Yuxing Lu\thanks{~~Equal contribution.}\thanks{~~Corresponding author.}\affA \affB,
  Yushuhong Lin\footnotemark[1]\affA,
  Jason Zhang\affA
  \\[4pt]
  \affA Georgia Institute of Technology \\
  \affB Peking University
  \\[4pt]
  \texttt{\{yxlu, ylin766, jzhang3497\}@gatech.edu }
  \\[6pt]
}

\newcommand{\affA}{\textsuperscript{$\spadesuit$}} 
\newcommand{\affB}{\textsuperscript{$\heartsuit$}} 

\begin{document}
\maketitle
\begin{abstract}
Large language models applied to clinical prediction exhibit case-level heterogeneity: simple cases yield consistent outputs, while complex cases produce divergent predictions under minor prompt changes. Existing single-agent strategies sample from one role-conditioned distribution, and multi-agent frameworks use fixed roles with flat majority voting, discarding the diagnostic signal in disagreement. We propose \textbf{CAMP} (\textbf{C}ase-\textbf{A}daptive \textbf{M}ulti-agent \textbf{P}anel), where an attending-physician agent dynamically assembles a specialist panel tailored to each case's diagnostic uncertainty. Each specialist evaluates candidates via three-valued voting (KEEP/REFUSE/NEUTRAL), enabling principled abstention outside one's expertise. A hybrid router directs each diagnosis through strong consensus, fallback to the attending physician's judgment, or evidence-based arbitration that weighs argument quality over vote counts. On diagnostic prediction and brief hospital course generation from MIMIC-IV across four LLM backbones, CAMP consistently outperforms strong baselines while consuming fewer tokens than most competing multi-agent methods, with voting records and arbitration traces offering transparent decision audits.
\end{abstract}

\section{Introduction}
When large language models are applied to clinical prediction tasks such as diagnosis prediction, their outputs exhibit notable sensitivity to prompt framing, specialist perspective, and the particular reasoning path during inference~\citep{hager2024evaluation,peng2023study}. This instability, however, is not uniform across cases. Some cases yield consistent predictions regardless of configuration, while others, particularly those involving overlapping conditions or ambiguous evidence, produce divergent outputs under minor changes to the inference setup. We refer to this phenomenon as \emph{case-level heterogeneity} in prediction, and argue that it reveals a structural mismatch between static inference strategies and the heterogeneous reasoning demands of individual cases.
 
Existing approaches address fragments of this challenge but do not resolve the gap. Prompting strategies such as chain-of-thought~\citep{wei2022chain} and self-consistency decoding~\citep{wang2022self} reduce output variance by marginalizing over sampled reasoning paths, yet all paths are drawn from a single role-conditioned distribution. The resulting diversity reflects sampling noise rather than genuinely distinct specialist perspectives, and there is no mechanism to adapt which expertise is invoked on a per-case basis~\citep{wu2025chain}. Multi-agent frameworks take a step further by instantiating diverse agents, but typically assign fixed roles across all cases and resolve disagreement through flat majority voting, which discards the diagnostic signal carried by the disagreement itself~\citep{tran2025multi}. No existing method jointly offers (1)~\emph{case-adaptive agent composition} and (2)~\emph{structured disagreement resolution} that leverages specialist reasoning.

These two capabilities mirror how clinical reasoning operates in practice~\citep{gruppen2011clinical}. When facing a complex or uncertain case, an attending physician does not consult a fixed panel of specialists. Instead, the choice of whom to consult is driven by the specific clinical presentation: which organ systems are involved, which findings are ambiguous, and where diagnostic uncertainty is greatest. Each consulted specialist independently evaluates the evidence and renders a judgment, endorsing, rejecting, or abstaining on each candidate diagnosis. When specialists largely agree, the attending physician accepts their consensus. When they disagree, however, the resolution is not a simple vote count: the attending physician examines each specialist's reasoning, weighs the strength of competing arguments, and renders a final judgment informed by the full body of evidence~\citep{taberna2020multidisciplinary}. This combination of adaptive expertise assembly and hierarchical arbitration suggests that a faithful computational model of the consultation process can address the limitations current methods share.

Building on this analysis, we propose \textbf{C}ase-\textbf{A}daptive \textbf{M}ulti-agent \textbf{P}anel (\textsc{\textbf{CAMP}}), a multi-agent architecture for clinical prediction. Given a clinical context, an attending-physician agent extracts structured evidence, generates candidate diagnoses, and dynamically assembles a specialist panel with case-adaptive role-conditioned prompts. Each specialist independently evaluates every candidate diagnosis through a structured three-valued vote. When specialists disagree, the attending physician performs informed arbitration: examining each specialist's reasoning and rendering a final decision that weighs argument quality rather than simply counting votes. We evaluate \textsc{CAMP} on two complementary tasks derived from the MIMIC-IV dataset. \textsc{CAMP} consistently outperforms strong baselines on both tasks, while offering higher interpretability through its explicit voting records and arbitration traces. Our contributions are as follows:
\begin{enumerate}[leftmargin=*,itemsep=2pt,topsep=4pt]
 
\item We propose \textsc{CAMP}, a multi-agent clinical prediction framework that combines case-adaptive specialist assembly with hierarchical arbitration, dynamically tailoring both the panel composition and the resolution strategy.
 
\item We introduce a structured three-valued voting mechanism (\textsc{Keep}/\textsc{Refuse}/\textsc{Neutral}) that allows specialists to abstain under genuine ambiguity, and show that this yields more informative judgment signals than binary voting.
 
\item We validate \textsc{CAMP} on diagnostic prediction and BHC generation tasks derived from MIMIC-IV, demonstrating consistent improvements over strong baselines.
 
\end{enumerate}

\section{Related Works}
 
\subsection{LLMs for Clinical Prediction}
 
Large language models have been increasingly applied to clinical prediction tasks, including diagnosis prediction~\citep{lu-etal-2024-clinicalrag}, clinical note summarization~\citep{madzime2024enhanced}, and brief hospital course generation~\citep{small2025evaluating}. These efforts have primarily focused on model selection, domain-adaptive pretraining, and task-specific fine-tuning~\citep{shool2025systematic}. More recently, inference-time reasoning strategies have been explored to improve prediction quality without modifying model parameters. Chain-of-thought prompting~\citep{wei2022chain} encourages step-by-step clinical reasoning, while self-consistency decoding~\citep{wang2022self} marginalizes over multiple sampled reasoning paths to reduce output variance. Medically tailored prompting strategies, such as those employed in Med-PaLM~\citep{singhal2023large} and other clinical LLM systems~\citep{lu2023medkpl}, further incorporate domain-specific heuristics to guide reasoning. However, all of these strategies draw reasoning paths from a single role-conditioned distribution, meaning that the resulting diversity reflects sampling variability rather than genuinely distinct specialist perspectives. Crucially, none provides a mechanism to adapt which expertise is invoked on a per-case basis, leaving the case-level heterogeneity problem unaddressed.
 
\subsection{Multi-Agent Inference}
 
Multi-agent frameworks improve LLM reasoning by instantiating multiple agents with distinct roles or perspectives. \citet{du2024improving} shows that inter-agent critique enhances factual accuracy, and subsequent work explores divergent thinking~\citep{liang2024encouraging}, role-differentiated debate~\citep{quan2025towards, lu2026dytopo}, and medical multi-disciplinary team simulation~\citep{han2025multi}. However, existing frameworks share 2 limitations. First, agent roles are fixed across all inputs, applying the same panel regardless of case-specific reasoning demands. Second, disagreement is typically resolved through flat majority voting or debate aimed at convergence, neither of which leverages the \emph{content} of specialist reasoning to inform the final decision. Adaptive inference methods such as Mixture-of-Experts~\citep{mu2025comprehensive} offer a conceptual parallel by routing inputs to specialized parameter subsets, but operate at the model level rather than at the level of agent roles and reasoning. 

\textsc{CAMP} differs from prior multi-agent work in 3 respects. It performs case-adaptive routing at agent level, assembling a specialist panel tailored to each input. It introduces a structured three-valued voting mechanism that captures richer judgment signals than binary or ranked outputs. And it resolves disagreement through hierarchical arbitration, in which an attending-physician agent examines specialist rationales and renders informed decisions, rather than reducing disagreement to a vote count.

\begin{figure*}
    \centering
    \includegraphics[width=\linewidth]{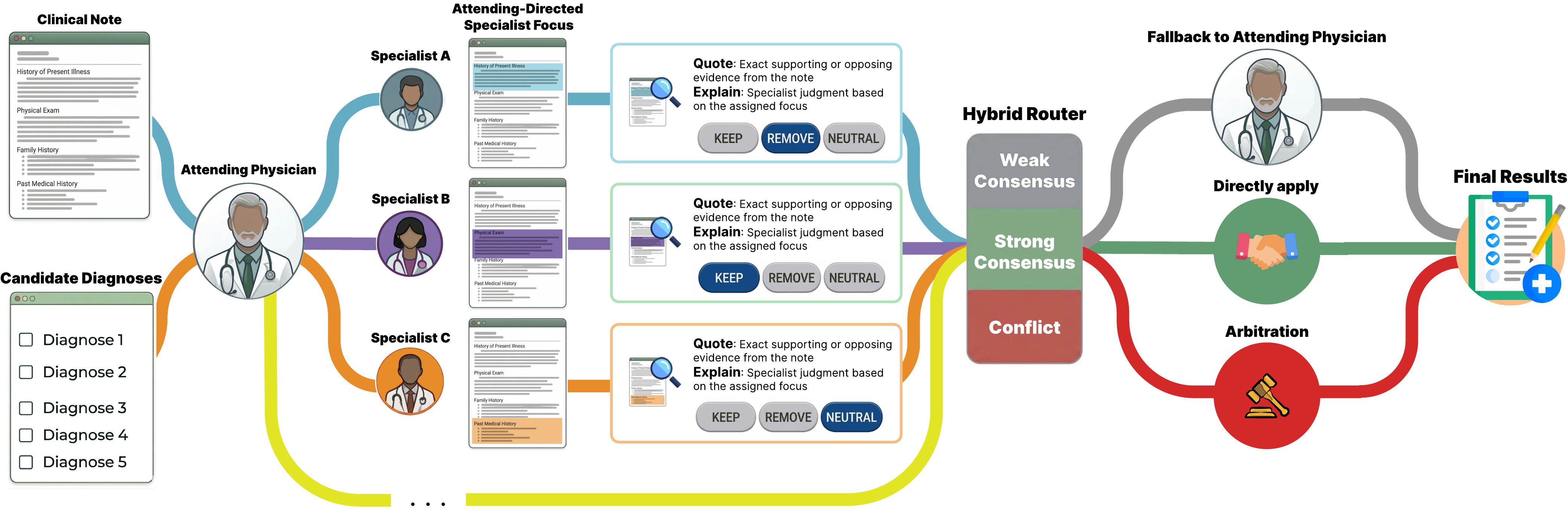}
    \caption{\textbf{Overview of the CAMP framework.} Given a clinical note and candidate diagnoses, the attending physician first renders an initial diagnostic judgment, then assembles a case-adaptive specialist panel with directed focus areas. Each specialist quotes evidence from the note and casts a three-valued vote. A hybrid router directs each diagnosis to one of three resolution paths: strong consensus decisions are applied directly, weak consensus falls back to the attending physician's initial judgment, and conflicts are escalated for evidence-based arbitration.}
    \label{fig:framework}
\end{figure*}

\section{Problem Setting}
\label{sec:problem}
 
We consider clinical prediction tasks in which a model receives a clinical context~$x$ and generate the correct answer $\mathcal{Y}$. We observe empirically that prediction consistency under different role-conditioned prompts varies across cases: many yield unanimous agreement while others produce divergent predictions under prompt changes, a phenomenon we term \emph{case-level heterogeneity} (Appendix~\ref{app:case_studies}).
This motivates 3 design requirements:
\begin{itemize}[leftmargin=*,itemsep=2pt,topsep=4pt]
    \item \textbf{R1} (\emph{Case-Adaptive Composition}). The specialist panel should vary across cases, tailored to each patient's sources of diagnostic uncertainty.
    \item \textbf{R2} (\emph{Gated Resolution}). Clear consensus should be accepted directly; expertise gaps should fall back to a generalist judgment; only genuine conflicts should trigger full arbitration.
    \item \textbf{R3} (\emph{Evidence-Informed Arbitration}). Conflicts should be resolved by reasoning over specialist arguments, not by vote counting.
\end{itemize}
These mirror clinical consultation practice~\citep{gruppen2011clinical,taberna2020multidisciplinary}, where an attending physician adapts whom to consult, accepts consensus, falls back on generalist judgment when specialists lack relevant expertise, and arbitrates disagreements by weighing argument quality.

\section{Methods}

\subsection{Overview}
\textsc{CAMP} consists of an attending-physician agent that orchestrates a dynamically assembled panel of specialist agents through four stages: (1)~initial assessment, (2)~case-adaptive panel assembly, (3)~specialist deliberation, and (4)~hybrid resolution. Figure~\ref{fig:framework} illustrates the overall architecture, and Algorithm~\ref{alg:camp} provides the complete pseudocode.
 
\subsection{Initial Assessment and Panel Assembly}
\label{sec:assembly}

\paragraph{Initial assessment.}
Given the clinical context~$x$ (organized into structured sections via rule-based preprocessing; see Appendix~\ref{app:data_preprocessing}) and the candidate diagnosis set~$\mathcal{Y}$, the attending-physician agent first performs an initial diagnostic assessment:
\begin{equation}
\small
    \{\hat{y}^{0}_{j}\}_{j=1}^{n} = \mathcal{M}_{\text{att}}(x,\,\mathcal{Y}),\quad \hat{y}^{0}_{j}\in\{\textsc{Accept},\,\textsc{Reject}\}.
\end{equation}
This initial judgment serves as a low-cost fallback for cases where the specialist panel lacks sufficient expertise (Section~\ref{sec:resolution}, Path~2). The attending physician also identifies the case's key clinical dimensions: the primary organ systems involved, the ambiguous findings, and the principal sources of diagnostic uncertainty, which inform the subsequent panel composition.

\paragraph{Case-adaptive panel assembly.}
Based on its analysis of the clinical context, the attending physician assembles a panel of $k$ specialist agents:
\begin{equation}
\small
    \mathcal{R} = \textsc{Assemble}(x) = \{r_1, \dots, r_k\},
\end{equation}
where each $r_i$ is a role-conditioned system prompt consisting of a specialty identity (e.g., ``neurologist'') and a case-specific focus directive that instructs the specialist which evidence to prioritize (e.g., \emph{``pay particular attention to the nonvisualization of the right vertebral artery and the wide-based gait''}). The panel composition varies across cases, satisfying \textbf{R1}: different clinical presentations yield different specialist configurations tailored to the relevant sources of diagnostic uncertainty. We adopt $k{=}3$ as the default panel size based on a sensitivity analysis (Section~\ref{sec:panel}).
 
\subsection{Specialist Deliberation}
\label{sec:voting}
Each specialist agent $\mathcal{A}_i$ evaluates all candidate diagnoses, producing a structured evaluation for every diagnosis:

\vspace{-10pt}
\begingroup
\small
\begin{align}
    \{(\hat{v}_{ij},\; q_{ij},\; c_{ij},\; &s_{ij})\}_{j=1}^{n} = \mathcal{M}(x, r_i, \mathcal{Y}), \nonumber \\
    \hat{v}_{ij} &\in \{\textsc{Keep}, \textsc{Refuse}, \textsc{Neutral}\}.
\end{align}
\endgroup
where $\hat{v}_{ij}$ is the specialist's vote (\textsc{Keep} endorses $d_j$, \textsc{Refuse} rejects it, \textsc{Neutral} indicates insufficient evidence to judge), $q_{ij}$ is an evidence quote extracted from the clinical note, $c_{ij} \in [0,1]$ is a confidence score, and $s_{ij}$ is a natural-language rationale grounded in the quoted evidence.
Across all $k$ specialists, this produces a vote matrix $\mathbf{V} \in \{\textsc{K}, \textsc{R}, \textsc{N}\}^{k \times n}$ and corresponding evidences.
 
The three-valued voting scheme is a deliberate design choice. Binary Keep/Refuse forces agents to commit even under genuine ambiguity, which introduces noise when a diagnosis falls outside a specialist's domain. The \textsc{Neutral} option allows agents to abstain in such cases: for instance, when an addiction specialist is asked to evaluate ``left ICA stenosis,'' a vascular diagnosis entirely outside the scope of addiction medicine, \textsc{Neutral} is the only appropriate response (see Case~10330554-DS-13 in Appendix~\ref{app:case_studies}). A high proportion of \textsc{Neutral} votes on a given diagnosis is itself an informative signal, it indicates that the current panel lacks collective expertise on that diagnosis, triggering the fallback path described below.

\subsection{Hybrid Resolution}
\label{sec:resolution}

Given the vote matrix $\mathbf{V}$, we resolve each candidate diagnosis $d_j$ independently. Let $k_j$, $r_j$, and $n_j$ denote the number of \textsc{Keep}, \textsc{Refuse}, and \textsc{Neutral} votes for~$d_j$:

\vspace{-10pt}
{\small
\begin{align}
    k_j &= \Sigma_{i=1}^{k} \mathds{1}[\hat{v}_{ij} = \textsc{K}], \nonumber \\
    r_j &= \Sigma_{i=1}^{k} \mathds{1}[\hat{v}_{ij} = \textsc{R}], \\
    n_j &= \Sigma_{i=1}^{k} \mathds{1}[\hat{v}_{ij} = \textsc{N}]. \nonumber
\end{align}}

A hybrid router directs each diagnosis through one of three resolution paths, satisfying \textbf{R2} and \textbf{R3}:

\paragraph{Path~1: Strong Consensus.}
If all non-abstaining specialists agree and form a strict majority:
\begin{equation}
\small
    \hat{y}_j = 
    \begin{cases}
        \textsc{Accept}  & \text{if } r_j = 0 \text{ and } k_j > n_j, \\
        \textsc{Reject}  & \text{if } k_j = 0 \text{ and } r_j > n_j.
    \end{cases}
\end{equation}
When specialists who express an opinion are unanimous and more than those who abstain, CAMP directly applies their consensus without further steps.

\paragraph{Path~2: Weak Consensus.}
If there is only 1 type of non-abstaining vote but abstentions dominate:
\begin{equation}
\small
    \hat{y}_j = \hat{y}^{0}_{j} \quad\text{if}\;\; k_j \cdot r_j = 0 \;\;\text{and}\;\; \max(k_j,\, r_j) \leq n_j.
\end{equation}
This covers cases where a majority of specialists voted \textsc{Neutral} (with the remainder agreeing), or where all specialists abstained ($k_j = r_j = 0$). In both scenarios, the panel lacks sufficient collective expertise on~$d_j$. Rather than invoking the arbitration procedure, the system falls back to the attending physician's initial judgment~$\hat{y}^{0}_{j}$: when the attending physician's generalist assessment formed during the initial case analysis is a more reliable signal than a poorly informed panel vote.

\paragraph{Path~3: Conflict.}
When both \textsc{Keep} and \textsc{Refuse} votes coexist ($k_j > 0$ and $r_j > 0$), the specialists have reached a genuine disagreement. The attending physician performs evidence-based arbitration:
\begin{equation}
\small
    \hat{y}_j = \textsc{Arbitrate}\!\left(x,\; d_j,\; \{(r_i, \hat{v}_{ij}, q_{ij}, c_{ij}, s_{ij})\}_{i=1}^{k}\right)
\end{equation}
The attending physician examines the original clinical context, the candidate diagnosis, and every specialist's vote together with its evidence quote, confidence, and rationale, then renders a final \textsc{Accept} or \textsc{Reject} judgment by weighing argument quality rather than simply counting votes (see Section~\ref{sec:conflict} for cases where the arbitrator overrides the majority by identifying stronger evidence in the minority's rationale).
 
\paragraph{Final Output.}
The predicted diagnosis set is:
\begin{equation}
\small
    \hat{\mathcal{Y}} = \{d_j \in \mathcal{Y} \mid \hat{y}_j = \textsc{Accept}\}
\end{equation}
This hierarchical design ensures computational efficiency, as the attending-physician arbitration call is reserved for genuinely contested diagnoses, while straightforward and expertise-gap cases are resolved without additional LLM calls.

\subsection{Adaptation to Text Generation}
\label{sec:bhc}
 
For the brief hospital course (BHC) generation task, the attending-physician agent generates the narrative by synthesizing the accepted diagnoses with the original clinical notes: $\text{BHC} = \mathcal{M}_{\text{attend}}(x,\, \hat{\mathcal{Y}})$.
The predicted diagnoses serve as a structured clinical summary that organizes the generation around the panel's verified diagnostic conclusions, grounding the BHC in deliberative reasoning rather than generating directly from raw notes alone.
 
\section{Experiments}
We evaluate CAMP on two tasks: diagnostic prediction and brief hospital course (BHC) generation, both derived from MIMIC-IV-Ext-BHC dataset~\citep{aali2025dataset}. The dataset provides 270,033 preprocessed discharge summary-BHC pairs from Beth Israel Deaconess Medical Center (2008-2019), derived from MIMIC-IV-Note~\citep{johnson2023mimic}.

\subsection{Dataset}

\paragraph{Diagnostic Prediction.}
We frame diagnosis prediction as a multi-label selection task. From each discharge summary, individual diagnoses are extracted via rule-based normalization and LLM-assisted refinement (Appendix~\ref{app:data_preprocessing} and~\ref{app:task_construction}). Each sample consists of a masked clinical note and a shuffled candidate set containing both correct diagnoses and distractors. We evaluate on 2,000 randomly sampled instances with a fixed seed for reproducibility.

\paragraph{BHC Generation.}
Given the same masked note and the accepted diagnoses~$\hat{\mathcal{Y}}$ from the diagnostic prediction pipeline, the model generates a brief hospital course narrative. This task tests whether structured diagnostic reasoning from CAMP improves downstream generation quality.

\subsection{LLM Backbones}
We evaluate on three LLM backbones: Llama-3.1-70B~\citep{grattafiori2024llama}, Gemma-3-27B~\citep{team2024gemma}, and GPT-OSS-20B~\citep{agarwal2025gpt}. All LLM calls use temperature~0 to ensure reproducible outputs. To demonstrate robustness, we additionally report CAMP and Single Agent results on GPT-5.4~\citep{openai2026gpt54}; full baseline comparisons on this backbone are left to future work due to cost constraints.
 
\subsection{Metrics}
For diagnostic prediction we report Macro F1 (per-diagnosis F1 averaged across all candidates) and Perfect Rate (percentage of instances whose predicted set exactly matches the ground truth). For BHC generation, all method outputs are ranked by an LLM judge (based on GPT-5-mini \citep{openai2024gpt5mini}, Appendix~\ref{app:prompt_comparison}), with lower average ranks indicating better performance; seven supplementary dimensions are reported in Appendix~\ref{app:bhc_additional_metrics}.
 
\begin{table*}[t]
\centering
\small
\setlength{\tabcolsep}{2pt}
\caption{Performance comparison across methods and LLM backbones. Best results within each model are bold.}
\label{tab:main_results}
\begin{tabular}{llccccc}
\toprule
\rowcolor{blue}
\camphead{Model} & \camphead{Method} 
& \multicolumn{2}{c}{\camphead{MIMIC-IV Diagnosis $\uparrow$}} 
& \multicolumn{3}{c}{\camphead{BHC Generation $\downarrow$}} \\
\cmidrule(lr){3-4} \cmidrule(lr){5-7}
& 
& \textbf{Macro F1} & \textbf{Perfect Rate} 
& \textbf{\shortstack[c]{Reasoning}}
& \textbf{\shortstack[c]{Readability}}
& \textbf{\shortstack[c]{Utility}} \\
\midrule

\multirow{8}{*}{\shortstack[l]{Llama-3.1-70B\\ \citep{grattafiori2024llama}}}
& Single Agent & 78.68 & 44.25 & 3.24 & 3.61 & 3.67 \\
& Self-Consistency~\citep{wangself} & 78.91 & 45.05 & 3.78 & 4.06 & 4.18 \\
& Chain-of-Thought~\citep{wei2022chain} & 78.99 & 45.55 & 3.47 & 3.80 & 3.92 \\
& Majority Voting~\citep{kaesberg2025voting} & 85.84 & 66.55 & 3.43 & 3.53 & 3.55 \\
& LLM-as-a-Judge~\citep{zheng2023judging} & 83.13 & 57.40 & 4.02 & 4.16 & 4.22 \\
& Devil's Advocate~\citep{ke2024mitigating} & 83.06 & 57.20 & 4.49 & 4.65 & 4.80 \\
& MedAgents~\citep{tang2024medagents} & 82.77 & 56.70 & 4.08 & 4.18 & 4.20 \\
& \cellcolor{bluelight}\campmetric{CAMP (Ours)} & \cellcolor{bluelight}\campmetric{87.50} & \cellcolor{bluelight}\campmetric{67.10} & \cellcolor{bluelight}\campmetric{2.49} & \cellcolor{bluelight}\campmetric{2.78} & \cellcolor{bluelight}\campmetric{2.73} \\
\midrule

\multirow{8}{*}{\shortstack[l]{GPT-OSS-20B\\ \citep{agarwal2025gpt}}}
& Single Agent & 75.78 & 60.45 & \campmetric{3.48} & 3.38 & 3.34 \\
& Self-Consistency~\citep{wangself} & 76.38 & 60.75 & 3.52 & 3.33 & 3.28 \\
& Chain-of-Thought~\citep{wei2022chain} & 76.54 & 60.50 & 3.63 & 3.44 & 3.43 \\
& Majority Voting~\citep{kaesberg2025voting} & 74.00 & 58.60 & 3.93 & 3.75 & 3.81 \\
& LLM-as-a-Judge~\citep{zheng2023judging} & 75.98 & 47.92 & 6.02 & 6.25 & 6.34 \\
& Devil's Advocate~\citep{ke2024mitigating} & 75.56 & 47.47 & 6.00 & 6.37 & 6.30 \\
& MedAgents~\citep{tang2024medagents} & 75.54 & 47.22 & 5.87 & 6.28 & 6.26 \\
& \cellcolor{bluelight}\campmetric{CAMP (Ours)} & \cellcolor{bluelight}\campmetric{82.20} & \cellcolor{bluelight}\campmetric{63.15} & \cellcolor{bluelight}3.55 & \cellcolor{bluelight}\campmetric{3.20} & \cellcolor{bluelight}\campmetric{3.24} \\
\midrule

\multirow{8}{*}{\shortstack[l]{Gemma-3-27B\\ \citep{team2024gemma}}}
& Single Agent & 81.12 & 49.60 & 4.11 & 4.30 & 4.30 \\
& Self-Consistency~\citep{wangself} & 81.02 & 49.45 & 4.30 & 4.51 & 4.43 \\
& Chain-of-Thought~\citep{wei2022chain} & 83.16 & 56.45 & 4.22 & 4.24 & 4.16 \\
& Majority Voting~\citep{kaesberg2025voting} & 78.17 & 58.05 & 5.78 & 6.00 & 5.97 \\
& LLM-as-a-Judge~\citep{zheng2023judging} & 82.22 & 56.15 & 4.68 & 4.68 & 4.76 \\
& Devil's Advocate~\citep{ke2024mitigating} & \campmetric{85.56} & \campmetric{65.50} & 4.49 & 4.24 & 4.19 \\
& MedAgents~\citep{tang2024medagents} & 85.34 & 65.20 & 5.14 & 5.05 & 5.08 \\
& \cellcolor{bluelight}\campmetric{CAMP (Ours)} & \cellcolor{bluelight}85.10 & \cellcolor{bluelight}64.20 & \cellcolor{bluelight}\campmetric{3.30} & \cellcolor{bluelight}\campmetric{2.97} & \cellcolor{bluelight}\campmetric{3.11} \\
\midrule

\multirow{2}{*}{\shortstack[l]{GPT-5.4\\ \citep{openai2026gpt54}}}
& Single Agent & 89.95 & 74.65 & 1.7 & 1.7 & 1.8 \\
& \cellcolor{bluelight}\campmetric{CAMP (Ours)} & \cellcolor{bluelight}\campmetric{91.31} & \cellcolor{bluelight}\campmetric{75.65} & \cellcolor{bluelight}\campmetric{1.3} & \cellcolor{bluelight}\campmetric{1.3} & \cellcolor{bluelight}\campmetric{1.2} \\
\bottomrule
\end{tabular}
\end{table*}

\section{Results}
We organize our evaluation around five research questions that progressively examine CAMP from overall effectiveness to component-level analysis. We first assess whether CAMP outperforms strong baselines on diagnostic prediction and BHC generation across different LLM backbones \textbf{(RQ1)}. We then trace the pipeline to examine each stage. At the panel assembly stage, we ask whether the specialists selected by the attending-physician agent correspond to the clinical services the patient actually encountered \textbf{(RQ2)}. At the voting stage, we investigate how the number of specialist agents affects prediction quality, confirming the value of multi-agent deliberation over single-agent inference \textbf{(RQ3)}. At the aggregation stage, we present case studies in which the majority of specialists voted to reject a diagnosis that was in fact correct; we show that the attending-physician arbitrator can recover these diagnoses by reasoning over the minority's supporting rationales, validating the hierarchical arbitration mechanism \textbf{(RQ4)}. Finally, we analyze token consumption to verify that CAMP's performance gains do not come at prohibitive computational cost \textbf{(RQ5)}.

\subsection{Overall Comparison (RQ1)}
\label{sec:rq1}
We compare CAMP against 7 baselines spanning three categories. \emph{Single-agent methods}: (1)~Single Agent, which prompts the LLM once without additional reasoning structure; and (2)~Chain-of-Thought~\citep{wei2022chain}, which elicits step-by-step reasoning. \emph{Multi-agent voting methods}: (3)~Self-Consistency~\citep{wang2022self}, which samples ten parallel reasoning paths and aggregates via majority vote; (4)~Majority Voting~\citep{kaesberg2025voting}, which runs multiple agents with identical prompts; and (5)~MedAgents~\citep{tang2024medagents}, which assigns fixed medical specialties to agents. \emph{Adjudication-based methods}: (6)~LLM-as-a-Judge~\citep{zheng2023judging}, which uses a separate LLM call to adjudicate among candidate predictions; and (7)~Devil's Advocate~\citep{ke2024mitigating}, which introduces a critic agent that challenges an initial prediction. All methods use the same LLM backbone within each model group.

\paragraph{Diagnostic Prediction.}
Table~\ref{tab:main_results} reports results across four LLM backbones. CAMP ranks first on three of four backbones and remains competitive on the fourth. The gain is largest on GPT-OSS-20B, where CAMP outperforms the next-best method by over five Macro~F1 points (82.20 vs.\ 76.54) and raises Perfect Rate from 60.75 to 63.15. On this backbone, all other multi-agent methods either underperform or only marginally improve over Single Agent, suggesting that smaller models may benefit most from structured role assignment. On Llama-3.1-70B, CAMP achieves the highest scores in both metrics (87.50 / 67.10). On Gemma-3-27B, CAMP (85.10) performs comparably to Devil's Advocate (85.56) and MedAgents (85.34), indicating that the advantage of case-adaptive composition varies across model architectures. On GPT-5.4, CAMP improves over Single Agent on both metrics (91.31 / 75.65 vs.\ 89.95 / 74.65).

Two cross-cutting observations worth attention. First, multi-agent strategies do not uniformly help: Majority Voting on GPT-OSS-20B (74.00) falls below Single Agent (75.78), and adjudicator-based methods show inconsistent gains across backbones. Second, among all methods tested, CAMP is the only one that consistently ranks in the top two across all four backbones on Macro~F1.

\paragraph{BHC Generation.}
Table~\ref{tab:main_results} (right) reports BHC generation results. Each dimension is scored by pooling and ranking all method outputs, so lower values indicate better performance (supplementary dimensions in Appendix~\ref{app:bhc_additional_metrics}). On Llama-3.1-70B, CAMP achieves the best rank across all three dimensions (2.49, 2.78, 2.73), outperforming Majority Voting, the next-best baseline, by a clear margin. Most multi-agent baselines receive worse ranks than Single Agent, which scores 3.24 / 3.61 / 3.67; Devil's Advocate receives the worst ranks across all dimensions (4.49, 4.65, 4.80), suggesting that adversarial critique without structured coordination introduces noise into open-ended generation. Results on other backbones are broadly consistent, with CAMP achieving the best ranks across most dimensions (Appendix~\ref{app:bhc_additional_metrics}). These results indicate that CAMP's case-adaptive panel and hierarchical arbitration benefit not only discrete classification but also open-ended clinical text generation.

\begin{figure}
    \centering
    \includegraphics[width=\linewidth]{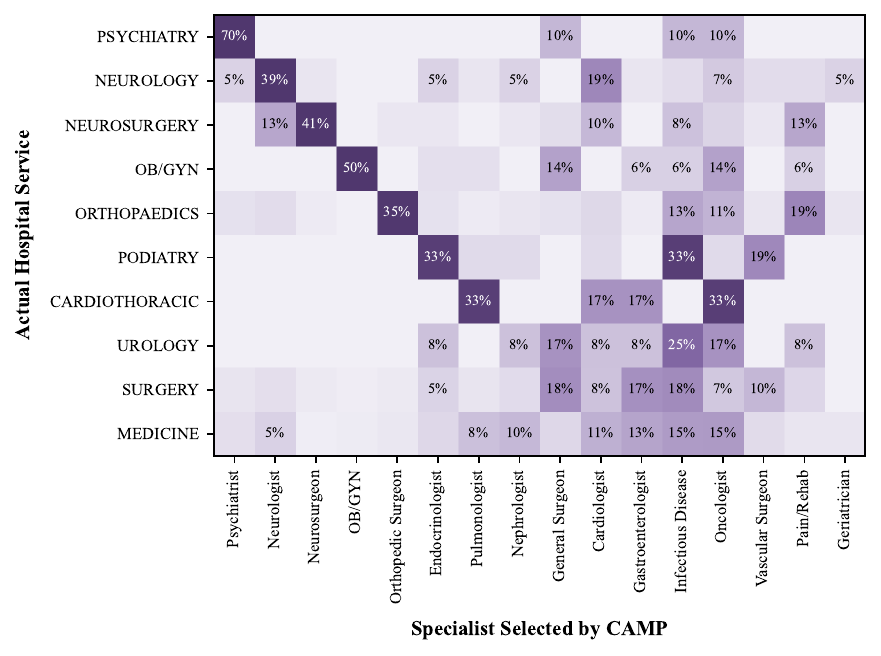}
    \caption{Alignment between specialists selected by CAMP and patients' actual hospital services. Each row is normalized to sum to 100\%. Specialist-specific services show strong concentration on the matching specialist, while general services exhibit broader distributions.}
    \label{fig:specialist_service}
\end{figure}

\subsection{Specialist--Service Alignment (RQ2)}
To validate that CAMP's case-adaptive panel assembly selects clinically relevant specialists, we compare the specialists chosen by the attending-physician agent against the hospital services each patient was actually admitted to. Figure~\ref{fig:specialist_service} shows the row-normalized distribution of specialist assignments across services, with rows and columns arranged to highlight diagonal correspondence.

The heatmap reveals a pronounced diagonal pattern for specialty-specific services: psychiatry patients receive a psychiatrist in 70\% of assignments, OB/GYN patients an OB/GYN specialist in 50\%, and neurosurgery, neurology, and orthopaedics patients receive the matching specialist in 35-41\% of cases. These concentrations indicate that the attending-physician agent identifies the relevant primary expertise without access to the service label. While this alignment does not directly establish a causal link to prediction quality, it provides evidence that the panel assembly mechanism selects clinically plausible specialists.

Off-diagonal assignments are also clinically plausible. For example, podiatry patients frequently receive an endocrinologist (33\%), reflecting diabetic foot comorbidities, and cardiothoracic patients are often assigned an oncologist (33\%), consistent with the prevalence of thoracic malignancies. For general services such as Medicine and Surgery, no single specialist exceeds 18\%, which is expected given the diversity of conditions these services cover.

\subsection{Effect of Panel Size (RQ3)}
\label{sec:panel}

Table~\ref{tab:agent_metrics} and Appendix Figure~\ref{fig:agent_num} show how the number of specialist agents affects diagnostic prediction, evaluated on a 200-case subset using Llama-3.1-70B. All three metrics follow a consistent pattern: performance increases sharply from one to three agents, then plateaus with minor fluctuations as more agents are added. The gain from one to three agents is substantial across all metrics. F1 rises from 64.84 to 68.97 and Recall from 70.83 to 74.33. The precision improvement is the largest in both absolute and relative terms (62.42 to 67.50, +8.1\%), suggesting that adding specialists primarily helps filter out incorrect diagnoses that a single agent would erroneously endorse.

Beyond three agents, all metrics fluctuate within a narrow band without a clear upward trend. For instance, Macro F1 at $k=4$ drops to 67.37, recovers to 68.17 at $k=5$, and remains in a similar range through $k=7$. This plateau is consistent with a saturation effect: once the panel covers the major clinical perspectives relevant to a case, additional specialists contribute overlapping viewpoints rather than genuinely new diagnostic signal. These results justify our default choice of $k=3$ as the default panel size in all other experiments.

\begin{table}[t]
\centering
\small
\setlength{\tabcolsep}{4pt}
\caption{Diagnostic prediction performance as a function of specialist panel size, evaluated on a 200-case subset using Llama-3.1-70B.}
\label{tab:agent_metrics}
\begin{tabular}{cccc}
\toprule
\rowcolor{blue}
\camphead{\# Agents} & \camphead{Macro F1} & \camphead{Macro Precision} & \camphead{Macro Recall} \\
\midrule
1 & 64.84 & 62.42 & 70.83 \\
2 & 67.77 & 66.33 & 72.33 \\
\rowcolor{bluelight}
3 & \campmetric{68.97} & 67.50 & \campmetric{74.33} \\
4 & 67.37 & 67.00 & 71.33 \\
5 & 68.17 & \campmetric{67.83} & 72.33 \\
6 & 67.22 & 66.83 & 70.83 \\
7 & 68.20 & 67.58 & 72.33 \\
\bottomrule
\end{tabular}
\end{table}

\begin{figure}[t]
    \centering
    \includegraphics[width=\linewidth]{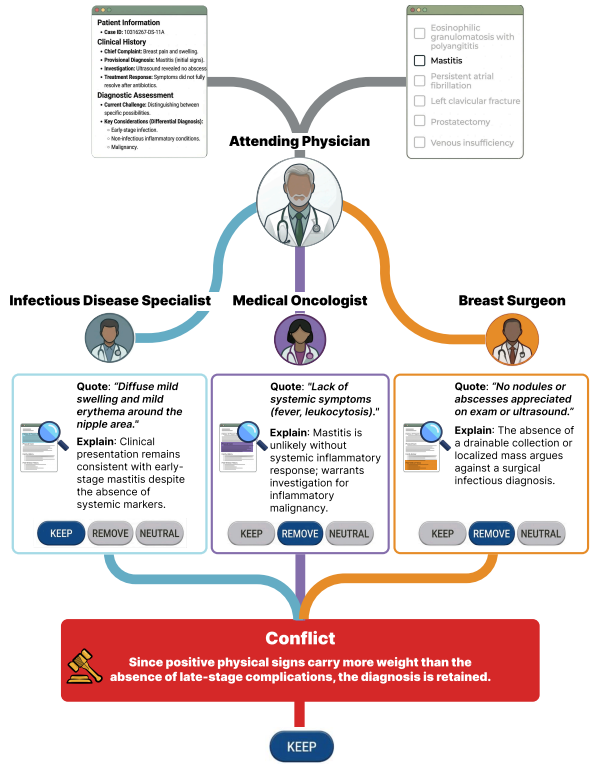}
    \caption{Conflict resolution via attending-physician arbitration. The attending physician overrides a 2-to-1 majority \textsc{Refuse} vote by weighing the quality of competing specialist rationales rather than counting votes.}
    \label{fig:conflict_case}
\end{figure}

\subsection{Conflict Resolution by Arbitration (RQ4)}
\label{sec:conflict}

A key advantage of CAMP is that genuine specialist disagreement is resolved through attending-physician arbitration rather than flat majority voting. This is particularly important when the majority opinion is driven by missing supporting evidence, while the minority identifies a more decisive positive clinical finding.

Figure~\ref{fig:conflict_case} shows a representative example from Case 10316267-DS-11. For the diagnosis \textit{Mastitis}, two specialists vote \textsc{Refuse} because the case lacks systemic inflammatory signs and imaging does not show an abscess or mass. In contrast, the infectious disease specialist votes \textsc{Keep} based on localized erythema and swelling, arguing that these are sufficient positive findings for early-stage mastitis. The attending physician examines both sides and accepts the diagnosis, reasoning that localized erythema and swelling constitute definitive positive findings, while the absence of abscess or fever reflects missing late-stage complications rather than evidence against the diagnosis. In other words, direct positive physical signs outweigh the absence of features that would only appear in more advanced presentations.

This example highlights why flat majority voting can be unreliable in clinically ambiguous settings. Specialist disagreement often reflects different evidential standards rather than simple model error. By explicitly weighing competing rationales, CAMP can recover correct diagnoses that would otherwise be discarded, while also producing an interpretable arbitration trace.

\begin{figure}[t]
    \centering
    \includegraphics[width=\linewidth]{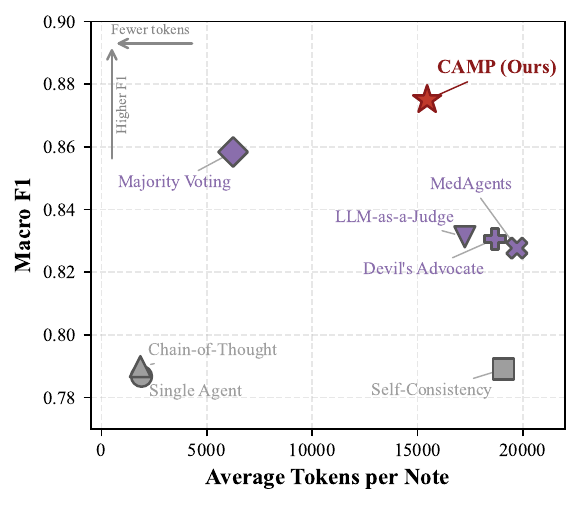}
    \caption{Token consumption vs.\ Macro F1. CAMP achieves the highest F1 while consuming fewer tokens than most of other multi-agent baselines.}
    \label{fig:token}
    \vspace{-5pt}
\end{figure}

\subsection{Token Consumption Analysis (RQ5)}
Figure~\ref{fig:token} plots average token consumption per note against Macro~F1 for all methods on Llama-3.1-70B. Simply scaling token budget through parallel sampling does not help: Self-Consistency consumes roughly 19K tokens yet barely improves over Single Agent ($\sim$1.9K tokens), because all paths are drawn from the same prompt distribution. Structured multi-agent methods (LLM-as-a-Judge, Devil's Advocate, MedAgents) achieve meaningful F1 gains but at 17-20K tokens per note. Majority Voting offers a favorable trade-off at $\sim$6.3K tokens with an F1 of 85.84, though it lacks any mechanism for handling disagreement beyond vote counting.
 
CAMP achieves the highest F1 (87.50) at ~15.5K tokens, fewer than all competing multi-agent methods except Majority Voting. Two design choices explain this: a compact three-specialist panel avoids the redundancy of larger fixed panels, and the gated resolution mechanism (Section~\ref{sec:resolution}) resolves consensus cases through direct vote counting, reserving the more expensive arbitration call for genuinely contested diagnoses. A detailed per-method breakdown is provided in Appendix~\ref{sec:cost_supp}.

\section{Conclusion}

We presented \textsc{CAMP}, a multi-agent architecture for clinical prediction that addresses case-level heterogeneity through three mechanisms: case-adaptive specialist assembly tailored to each patient's clinical profile, three-valued voting that allows specialists to abstain under genuine ambiguity, and a gated resolution procedure that routes each diagnosis through consensus, fallback, or evidence-based arbitration depending on the pattern of specialist agreement. Experiments on diagnostic prediction and BHC generation from MIMIC-IV show that CAMP achieves the best on diagnostic prediction and BHC generation tasks across evaluation metrics, and consumes fewer tokens than most competing multi-agent methods. Beyond accuracy, the explicit voting records and arbitration traces produced by CAMP offer a transparent decision audit that may support downstream clinical review.

\clearpage
\section*{Limitations}
 
All experiments use English-language notes from MIMIC-IV, a single-center U.S.\ dataset whose documentation style and coding conventions may not generalize to other institutions, healthcare systems, or languages. As the results illustrate, \textsc{CAMP}'s benefits diminish when the base model already achieves strong performance through alternative prompting strategies, and performance depends on model-specific prompt sensitivity that may limit reproducibility on untested LLMs. The BHC evaluation relies on LLM-based ranking rather than clinician judgment, which remains an important direction for future validation. Finally, although hierarchical aggregation limits unnecessary computation by reserving arbitration for contested diagnoses, multi-agent inference still requires several LLM calls per case, which may be impractical in latency-sensitive clinical workflows.

\section*{Ethical Considerations}
 
\textsc{CAMP} is a research prototype for studying multi-agent reasoning in clinical prediction and is not intended or validated for autonomous clinical decision-making; any deployment would require prospective evaluation, regulatory approval, and physician oversight. All experiments use de-identified MIMIC-IV records under the PhysioNet data use agreement, and no additional patient data were collected. Because the framework delegates reasoning to LLM-instantiated agents, demographic and socioeconomic biases encoded in the underlying model may propagate into panel assembly, voting, and arbitration; a formal bias audit is beyond the scope of this work but is an important direction for future study. The apparent interpretability of specialist votes and arbitration rationales should not be conflated with clinical validity, as LLM-generated reasoning can be fluent yet factually incorrect, risking undue anchoring if presented to clinicians without appropriate safeguards. Multi-agent inference also increases energy consumption relative to single-agent methods, and the cumulative environmental cost of large-scale deployment warrants consideration.

\bibliography{custom}

\clearpage
\appendix
\label{sec:appendix}

\section{CAMP Algorithm Pseudocode}
\label{sec:algorithm}

\begin{algorithm}[h]
\caption{Case-Adaptive Multi-Agent Panel}
\label{alg:camp}
\begin{algorithmic}[1]
\REQUIRE Clinical context $x$, candidate diagnoses $\mathcal{Y} = \{d_1, \dots, d_n\}$
\ENSURE Set of accepted diagnoses $\hat{\mathcal{Y}}$
\STATE \COMMENT{\textit{Stage 1: Initial Assessment and Panel Assembly (Sec.~\ref{sec:assembly})}}
\STATE $\{\hat{y}^{0}_{j}\}_{j=1}^{n} \leftarrow \mathcal{M}_{\text{att}}(x,\, \mathcal{Y})$ \COMMENT{Initial judgment}
\STATE $\mathcal{R} \leftarrow \textsc{Assemble}(x) = \{r_1, \dots, r_k\}$
\STATE \COMMENT{\textit{Stage 2: Specialist Deliberation (Sec.~\ref{sec:voting})}}
\FOR{$i = 1, \dots, k$}
    \STATE $\{(\hat{v}_{ij},\, q_{ij},\, c_{ij}, \, s_{ij})\}_{j=1}^{n} \leftarrow \mathcal{M}(x,\, r_i,\, \mathcal{Y})$
\ENDFOR
\STATE \COMMENT{\textit{Stage 3: Hybrid Resolution (Sec.~\ref{sec:resolution})}}
\STATE $\hat{\mathcal{Y}} \leftarrow \emptyset$
\FOR{$j = 1, \dots, n$}
    \STATE $k_j \leftarrow \sum_i \mathds{1}[\hat{v}_{ij} = \textsc{K}]$;\; $r_j \leftarrow \sum_i \mathds{1}[\hat{v}_{ij} = \textsc{R}]$;\; $n_j \leftarrow \sum_i \mathds{1}[\hat{v}_{ij} = \textsc{N}]$
    \IF{$r_j = 0$ \AND $k_j > n_j$}
    \STATE \COMMENT{Path 1: Strong consensus -- Accept}
        \STATE $\hat{\mathcal{Y}} \leftarrow \hat{\mathcal{Y}} \cup \{d_j\}$ 
    \ELSIF{$k_j = 0$ \AND $r_j > n_j$}
        \STATE \COMMENT{Path 1: Strong consensus -- Reject}
        \STATE \textbf{continue} 
    \ELSIF{$k_j \cdot r_j = 0$ \AND $\max(k_j, r_j) \leq n_j$}
    \IF{$\hat{y}^{0}_{j} = \textsc{Accept}$}
    \STATE \COMMENT{Path 2: Weak consensus -- Fallback}
        \STATE $\hat{\mathcal{Y}} \leftarrow \hat{\mathcal{Y}} \cup \{d_j\}$
    \ENDIF
    \ELSE
        \STATE $\hat{y}_j \leftarrow \textsc{Arb}(x,\, d_j,\, \{(r_i, \hat{v}_{ij}, q_{ij}, c_{ij}, s_{ij})\}_{i=1}^{k})$
        \COMMENT{Path 3: Conflict -- Arbitration}
        \IF{$\hat{y}_j = \textsc{Accept}$}
            \STATE $\hat{\mathcal{Y}} \leftarrow \hat{\mathcal{Y}} \cup \{d_j\}$
        \ENDIF
    \ENDIF
\ENDFOR
\RETURN $\hat{\mathcal{Y}}$
\end{algorithmic}
\end{algorithm}

Algorithm~\ref{alg:camp} presents the complete pseudocode for the CAMP pipeline. The procedure takes a clinical context as input and returns a set of accepted diagnoses through three sequential stages. In Stage~1, the attending-physician agent extracts structured evidence from the clinical note and assembles a case-adaptive specialist panel. In Stage~2, each specialist independently evaluates every candidate diagnosis using the three-valued voting scheme. In Stage~3, the hierarchical aggregation procedure resolves each diagnosis: unanimous or majority consensus leads to direct acceptance or rejection, while contested or ambiguous cases are escalated to the attending-physician agent for evidence-based arbitration.

\section{Data Preprocessing}
\label{app:data_preprocessing}
 
We extract clinical records from MIMIC-IV discharge summaries. Given that the \texttt{DISCHARGE DIAGNOSIS} section frequently contains complex formatting, nested sub-lists, and implicit groupings that resist direct parsing, we first employ an LLM-assisted diagnosis extraction technique. The model is instructed to refine the raw text into individual, verbatim diagnosis entries. These extracted terms serve both as the ground-truth labels for the current note and as candidate material for the global diagnosis pool used in distractor sampling.

Following the extraction of structured diagnoses, we apply a multi-stage filtering pipeline to ensure dataset quality. First, to maintain a manageable label space and avoid cases with overly diffuse diagnostic conclusions, we retain only those records with no more than three final diagnoses. Second, we utilize an LLM-based semantic filter to assess whether the discharge diagnoses are "directly or partially" recoverable from the history of present illness, past medical history, or medication-related content. Cases exhibiting excessive informational overlap are excluded to preserve the inferential integrity of the diagnostic task.

We then parse these tagged sections and retain only those that do not leak discharge-time information. Sections such as \texttt{DISCHARGE DIAGNOSIS}, \texttt{DISCHARGE INSTRUCTIONS}, \texttt{DISCHARGE CONDITION}, \texttt{DISCHARGE DISPOSITION}, \texttt{FOLLOWUP INSTRUCTIONS}, and \texttt{DISCHARGE MEDICATIONS} are masked from the model input. Conversely, admission-time and in-hospital sections, including demographics, vital signs, laboratory results, medications, and clinical notes are preserved. This masking mechanism ensures that the model must infer diagnoses from the cumulative clinical course rather than simply retrieving them from the discharge summary.

\begin{table}[t]
\centering
\small
\setlength{\tabcolsep}{1.5pt}
\caption{Descriptive statistics of the dataset.}
\label{tab:data_stats}
\begin{tabular}{lccccc}
\toprule
\rowcolor{blue}
\camphead{Metric} & \camphead{Mean} & \camphead{Std} & \camphead{Min} & \camphead{Median} & \camphead{Max} \\
\midrule
Options per case & 7.239 & 1.921 & 6 & 6 & 12 \\
\rowcolor{bluelight}
Correct labels per case & 1.505 & 0.692 & 1 & 1 & 3 \\
Distractors per case & 5.734 & 1.253 & 5 & 5 & 9 \\
\rowcolor{bluelight}
Input words & 621 & 382 & 46 & 537 & 3353 \\
Input characters & 3991 & 2447 & 325 & 3434 & 22391 \\
\rowcolor{bluelight}
Correct option ratio (\%) & 19.92 & 4.07 & 16.67 & 16.67 & 25.00 \\
\bottomrule
\end{tabular}
\end{table}

From the \texttt{DISCHARGE DIAGNOSIS} section of each note, we extract individual diagnoses through a multi-step normalization pipeline. First, section-level prefixes such as ``Primary:'', ``Secondary:'', and axis labels (e.g., ``Axis I:'') are stripped. The remaining text is split on newlines and inline delimiters (e.g., hyphens and numbered markers) to isolate individual entries. Each entry is further cleaned by removing bullet characters, trailing periods, placeholder tokens (e.g., ``stage \_\_\_''), and entries shorter than three characters. Comma-separated lists within a single entry are expanded into separate diagnoses. After normalization, entries that exceed a length threshold or contain residual multi-diagnosis patterns are re-split recursively. The union of all normalized diagnoses across the dataset forms a global \emph{diagnosis pool}, which serves as the source for both correct answers and candidate distractors in downstream task construction. Records that yield no valid diagnoses after normalization are excluded from the evaluation set.

\begin{table*}[t]
\centering
\small
\setlength{\tabcolsep}{3.8pt}
\caption{Supplementary BHC evaluation dimensions across LLM backbones. All scores are pooled ranks (lower is better). Best results within each model are bold.}
\label{tab:bhc_additional_metrics}
\begin{tabular}{llccccccc}
\toprule
\rowcolor{blue}
\camphead{Model} & \camphead{Method}
& \camphead{Correct.} 
& \camphead{Coverage} 
& \camphead{Coherence} 
& \camphead{Linkage} 
& \camphead{Priority} 
& \camphead{Tracking} 
& \camphead{Density} \\
\midrule

\multirow{8}{*}{\shortstack[l]{Llama-3.1-70B\\ \citep{grattafiori2024llama}}}
& Single Agent & 3.73 & 3.08 & 3.24 & 3.27 & 3.55 & 3.45 & 3.92 \\
& Self-Consistency & 4.00 & 3.78 & 3.88 & 3.88 & 3.98 & 3.84 & 4.00 \\
& Chain-of-Thought & 3.96 & 3.73 & 3.69 & 3.67 & 3.84 & 3.67 & 4.10 \\
& Majority Voting & 2.88 & 3.78 & 3.18 & 3.53 & 3.63 & 3.16 & 3.49 \\
& LLM-as-a-Judge & 3.65 & 4.80 & 3.63 & 3.71 & 4.39 & 3.82 & 3.94 \\
& Devil's Advocate & 3.98 & 4.57 & 4.00 & 4.31 & 4.45 & 4.37 & 4.59 \\
& MedAgents & 4.04 & 4.16 & 3.71 & 4.00 & 4.27 & 3.92 & 4.39 \\
& \cellcolor{bluelight}\textbf{CAMP (Ours)} & \cellcolor{bluelight}\textbf{2.08} & \cellcolor{bluelight}\textbf{2.16} & \cellcolor{bluelight}\textbf{2.49} & \cellcolor{bluelight}\textbf{2.20} & \cellcolor{bluelight}\textbf{2.59} & \cellcolor{bluelight}\textbf{2.53} & \cellcolor{bluelight}\textbf{2.82} \\
\midrule

\multirow{8}{*}{\shortstack[l]{GPT-OSS-20B\\ \citep{agarwal2025gpt}}}
& Single Agent & 3.27 & 3.39 & 3.38 & 3.38 & 3.29 & 3.36 & 3.22 \\
& Self-Consistency & 3.22 & \textbf{3.24} & \textbf{3.37} & 3.36 & 3.25 & 3.52 & 3.30 \\
& Chain-of-Thought & 3.30 & 3.36 & 3.48 & 3.48 & 3.55 & 3.49 & 3.45 \\
& Majority Voting & 3.84 & 3.82 & 3.79 & 3.87 & 3.83 & 3.78 & 3.88 \\
& LLM-as-a-Judge & 6.32 & 6.28 & 6.01 & 6.14 & 6.22 & 5.98 & 6.17 \\
& Devil's Advocate & 6.47 & 6.32 & 6.22 & 6.22 & 6.35 & 6.20 & 6.42 \\
& MedAgents & 6.44 & 6.23 & 6.07 & 6.26 & 6.29 & 6.06 & 6.44 \\
& \cellcolor{bluelight}\textbf{CAMP (Ours)} & \cellcolor{bluelight}\textbf{3.14} & \cellcolor{bluelight}3.36 & \cellcolor{bluelight}3.68 & \cellcolor{bluelight}\textbf{3.29} & \cellcolor{bluelight}\textbf{3.22} & \cellcolor{bluelight}\textbf{3.61} & \cellcolor{bluelight}\textbf{3.12} \\
\midrule

\multirow{8}{*}{\shortstack[l]{Gemma-3-27B\\ \citep{team2024gemma}}}
& Single Agent & 4.32 & 4.11 & 4.43 & 4.24 & 4.32 & 4.05 & 4.95 \\
& Self-Consistency & 4.68 & 4.32 & 4.11 & 4.41 & 4.27 & 4.51 & 4.68 \\
& Chain-of-Thought & 4.08 & 4.19 & 4.08 & 4.16 & 4.35 & 3.97 & 4.32 \\
& Majority Voting & 5.81 & 6.14 & 5.76 & 5.97 & 6.05 & 5.68 & 5.68 \\
& LLM-as-a-Judge & 4.76 & 4.95 & 4.62 & 4.73 & 4.68 & 4.97 & 4.57 \\
& Devil's Advocate & 4.62 & 4.51 & 4.38 & 4.30 & 4.24 & 4.16 & 4.16 \\
& MedAgents & 4.92 & 4.89 & 5.03 & 4.95 & 5.11 & 5.14 & 4.65 \\
& \cellcolor{bluelight}\textbf{CAMP (Ours)} & \cellcolor{bluelight}\textbf{2.81} & \cellcolor{bluelight}\textbf{2.89} & \cellcolor{bluelight}\textbf{3.59} & \cellcolor{bluelight}\textbf{3.24} & \cellcolor{bluelight}\textbf{2.97} & \cellcolor{bluelight}\textbf{3.51} & \cellcolor{bluelight}\textbf{3.00} \\
\midrule

\multirow{2}{*}{\shortstack[l]{GPT-5.4\\ \citep{openai2026gpt54}}}
& Single Agent & 1.6 & 1.7 & 1.7 & 1.7 & 1.7 & 1.7 & \campmetric{1.3} \\
& \cellcolor{bluelight}\textbf{CAMP (Ours)} & \cellcolor{bluelight}\campmetric{1.4} & \cellcolor{bluelight}\campmetric{1.3} & \cellcolor{bluelight}\campmetric{1.3} & \cellcolor{bluelight}\campmetric{1.3} & \cellcolor{bluelight}\campmetric{1.3} & \cellcolor{bluelight}\campmetric{1.3} & \cellcolor{bluelight}1.7 \\
\bottomrule
\end{tabular}
\end{table*}
 
\section{Diagnostic Task Construction}
\label{app:task_construction}
 
We frame diagnostic prediction as a multi-label selection task: given a masked clinical note and a curated set of candidate diagnoses, the model must identify all correct diagnoses while rejecting distractors. The task construction proceeds through three distinct phases.

\paragraph{Phase 1: Diagnosis-Aware Note Masking.}
Beyond section-level removal, we perform fine-grained masking to eliminate phrases within the remaining text that explicitly name or strongly reveal any known diagnoses. An LLM with tool-calling capabilities is invoked; for each diagnosis-revealing phrase identified, the model calls a structured \texttt{mask\_phrase} tool to extract the exact span. All identified spans are replaced with a placeholder token (``\_\_\_''), ordered by length to prevent partial-match artifacts. This step ensures that diagnoses cannot be trivially recovered from in-text mentions within otherwise retained sections, such as treatment notes referencing a condition by name.

\paragraph{Phase 2: Distractor Sampling and Filtering.}
Candidate distractors are sampled from the global diagnosis pool, excluding any labels assigned to the current record. To accommodate varying clinical complexities, we implement a dynamic candidate scaling strategy: for cases with 1, 2, or 3 correct diagnoses, the total number of candidate options is set to 6, 8, or 12, respectively. This proportional design maintains task difficulty while effectively differentiating the model's diagnostic screening capabilities across levels of complexity. To prevent near-duplicate contamination, an LLM-based filtering step removes candidates that are synonyms, abbreviations, or semantic paraphrases of the correct diagnoses, ensuring that distractors represent clearly distinct medical conditions. Filtered distractor lists are cached per note for reproducibility.

\paragraph{Phase 3: Question Assembly.}
The final candidate set is formed by concatenating the correct diagnoses with the filtered distractors, followed by shuffling with a fixed random seed. As a final safeguard, we perform an exact-match masking pass over the assembled note. Any substring matching a candidate option (case-insensitive, length $> 4$) is replaced with the placeholder token. This dynamic masking operates across both the main text and individual fields, ensuring that no candidate is recoverable through surface-level string matching. The resulting structure, comprising the masked clinical note, the shuffled candidate list, and the ground-truth label set, constitutes a single evaluation instance.

\section{Supplementary BHC Generation Evaluation}
\label{app:bhc_additional_metrics}

To provide a more comprehensive assessment of BHC generation quality beyond the three primary dimensions reported in the main text, we evaluate all methods on seven supplementary dimensions. Correctness measures whether the principal diagnoses are stated accurately. Coverage assesses the inclusion of major diagnoses, clinical events, and care decisions. Coherence captures the logical and temporal consistency of the generated narrative. Linkage evaluates whether diagnoses are explicitly connected to their corresponding treatments and follow-up actions. Priority reflects the emphasis placed on the most clinically significant problems relative to peripheral details. Tracking measures the consistency with which complications and outcomes are followed throughout the hospital course. Density captures the ratio of clinically useful information to unnecessary redundancy. All dimensions are scored through pooled ranking, with lower values indicating better performance.

Table~\ref{tab:bhc_additional_metrics} reports these results on Llama-3.1-70B. CAMP achieves the lowest (best) score on every supplementary dimension, demonstrating broad improvements across multiple aspects of note quality rather than gains confined to a single criterion. Compared to the Single Agent baseline, CAMP reduces Correctness from 3.73 to 2.08, Coverage from 3.08 to 2.16, Coherence from 3.24 to 2.49, Linkage from 3.27 to 2.20, Priority from 3.55 to 2.59, Tracking from 3.45 to 2.53, and Density from 3.92 to 2.82. Averaging across all seven dimensions, CAMP scores 2.41, compared with 3.38 for Majority Voting (the strongest competing baseline) and 3.46 for Single Agent. These results indicate that CAMP simultaneously improves factual faithfulness, clinical completeness, narrative organization, and information selection. Among comparison methods, Majority Voting is the most competitive overall, whereas LLM-as-a-Judge, Devil's Advocate, and MedAgents consistently underperform on most dimensions, suggesting that unstructured critique or fixed-role aggregation introduces noise that degrades open-ended generation quality.

\section{Prompt Templates}
\label{app:prompt_templates}
 
This section presents the prompt templates used in our experiments, covering the main diagnosis selection pipeline (Section~\ref{app:prompt_main}), comparison methods (Section~\ref{app:prompt_comparison}), and the downstream BHC generation and evaluation pipeline (Section~\ref{app:prompt_bhc}).
 
\subsection{Main Diagnosis Selection Pipeline}
\label{app:prompt_main}
 
\begin{tcolorbox}[promptbox, title=Single Baseline]
You are an attending physician reviewing a clinical note to determine discharge diagnoses.
 
\medskip
\textbf{Clinical Note:} \texttt{\{clinical\_note\}}
 
\medskip
\textbf{Candidate Diagnoses:} \texttt{\{options\_text\}}
 
\medskip
\textbf{Instructions:}
This is a multi-label task. Select only diagnoses supported by the inpatient course during this hospitalization. Do not include conditions mentioned only in past medical history unless they were actively managed during this admission. Do not include diagnoses explicitly negated in the note. For each selected diagnosis, output its exact 1-based index and brief supporting evidence.
\end{tcolorbox}
 
\begin{tcolorbox}[promptbox, title=CoT Baseline]
You are an attending physician reviewing a clinical note to determine discharge diagnoses.
 
\medskip
\textbf{Clinical Note:} \texttt{\{clinical\_note\}}
 
\medskip
\textbf{Candidate Diagnoses:} \texttt{\{options\_text\}}
 
\medskip
\textbf{Instructions:}
This is a multi-label task. Select only diagnoses supported by the inpatient course during this hospitalization. Exclude past-medical-history-only and explicitly negated diagnoses. Think step by step about the hospital course before answering. For each selected diagnosis, output its exact 1-based index and brief supporting evidence.
\end{tcolorbox}
 
\begin{tcolorbox}[promptbox, title=Confidence-Aware Initial Review]
You are an attending physician reviewing a clinical note to determine discharge diagnoses.
 
\medskip
\textbf{Clinical Note:} \texttt{\{clinical\_note\}}
 
\medskip
\textbf{Candidate Diagnoses:} \texttt{\{options\_text\}}
 
\medskip
\textbf{Instructions:}
This is a multi-label task. Select only diagnoses supported by the inpatient course during this hospitalization. Exclude past-medical-history-only and explicitly negated diagnoses. Think step by step about the hospital course before answering. For each selected diagnosis, output its exact 1-based index and a confidence label (\texttt{high} or \texttt{medium}); omit low-confidence diagnoses.
\end{tcolorbox}
 
\begin{tcolorbox}[promptbox, title=Orchestrator]
You are a chief physician coordinating a diagnostic consultation.
 
\medskip
\textbf{Clinical Note:} \texttt{\{clinical\_note\}}
 
\medskip
\textbf{Candidate Discharge Diagnoses:} \texttt{\{options\_text\}}
 
\medskip
\textbf{Task:}
First summarize the key clinical issues driving this admission in 1--2 sentences. Then recruit exactly three specialists whose expertise is most relevant to this case. For each specialist, provide a clinical role and a one-sentence focus describing the evidence that the specialist should prioritize.
\end{tcolorbox}
 
\begin{tcolorbox}[promptbox, title=Specialist Review]
You are a \texttt{\{role\}}.
 
\medskip
\textbf{Clinical Note:} \texttt{\{clinical\_note\}}
 
\medskip
\textbf{Candidate Diagnoses:} \texttt{\{candidates\_text\}}
 
\medskip
\textbf{Task:}
Review every candidate diagnosis from your specialist perspective. For each diagnosis, provide a decision (\texttt{KEEP}, \texttt{REMOVE}, or \texttt{NEUTRAL}), confidence, whether it is in scope, the level of evidence support, an evidence quote from the note, and brief reasoning. Only keep diagnoses supported by the inpatient course during this hospitalization; do not keep past-medical-history-only or explicitly negated diagnoses.
\end{tcolorbox}
 
\begin{tcolorbox}[promptbox, title=Arbitration]
You are an attending physician making final discharge-diagnosis decisions.
 
\medskip
\textbf{Clinical Note:} \texttt{\{clinical\_note\}}
 
\medskip
\textbf{Contested Diagnoses with Specialist Quotes, Reasoning, and Confidence:} \texttt{\{contested\_evidence\}}
 
\medskip
\textbf{Task:}
For each contested diagnosis, decide \texttt{INCLUDE} or \texttt{EXCLUDE} based on the clinical note and the collected evidence. Include only diagnoses supported by the inpatient course during this hospitalization. Exclude past-medical-history-only or explicitly negated diagnoses. Verify quoted evidence against the note and disregard unsupported or unverifiable claims.
\end{tcolorbox}
 
\subsection{Comparison Methods}
\label{app:prompt_comparison}
 
\begin{tcolorbox}[promptbox, title=Majority Vote]
You are Agent \texttt{\{agent\_id\}}. You must independently review the clinical note and decide for each candidate diagnosis.
 
\medskip
\textbf{Clinical Note:} \texttt{\{clinical\_note\}}
 
\medskip
\textbf{Candidate Diagnoses:} \texttt{\{options\_text\}}
 
\medskip
\textbf{Rules:}
This is a multi-label task. Only include diagnoses supported by the inpatient course during this hospitalization. Do not include past-medical-history-only diagnoses unless actively managed during this admission. Do not include diagnoses explicitly negated in the note. If evidence is absent or weak, choose \texttt{REMOVE}. Return one decision per candidate diagnosis using \texttt{KEEP} or \texttt{REMOVE}, together with a brief rationale.
\end{tcolorbox}
 
\begin{tcolorbox}[promptbox, title=MedAgents-Style Revote]
You are Agent \texttt{\{agent\_id\}}. You may revise your decisions after reading other agents' opinions.
 
\medskip
\textbf{Clinical Note:} \texttt{\{clinical\_note\}}
 
\medskip
\textbf{Candidate Diagnoses:} \texttt{\{options\_text\}}
 
\medskip
\textbf{Other Agents' Opinions:} \texttt{\{peer\_opinions\}}
 
\medskip
\textbf{Rules:}
Be evidence-based and do not conform for consensus. If peers provide direct evidence you missed, update accordingly. If peers' reasoning is weak or hallucinated, keep your prior decision. Maintain the same decision format, giving a \texttt{KEEP} or \texttt{REMOVE} judgment for each candidate diagnosis together with a brief rationale.
\end{tcolorbox}
 
\begin{tcolorbox}[promptbox, title=Devil's Advocate]
You are the Devil's Advocate in a diagnostic panel.
 
\medskip
\textbf{Clinical Note:} \texttt{\{clinical\_note\}}
 
\medskip
\textbf{Proposed Diagnoses:} \texttt{\{pool\_text\}}
 
\medskip
\textbf{Task:}
Challenge every proposed diagnosis. For each one, identify evidence that contradicts or fails to support its inclusion. If you genuinely cannot challenge it because the evidence is strong, vote \texttt{KEEP}. Vote \texttt{REMOVE} only when you can provide a specific counterargument grounded in the clinical note. Be rigorous but fair, and do not remove well-supported diagnoses.
\end{tcolorbox}

\subsection{BHC Generation and Evaluation}
\label{app:prompt_bhc}
 
\begin{tcolorbox}[promptbox, title=BHC Generation]
You are a hospitalist writing a Brief Hospital Course (BHC) for discharge documentation.
 
\medskip
\textbf{Source Note} (the original discharge-diagnosis section has been removed)\textbf{:} \texttt{\{clinical\_note\}}
 
\medskip
\textbf{Predicted Diagnoses:} \texttt{\{diagnosis\_list\}}
 
\medskip
\textbf{Task:}
Write a concise and clinically useful BHC grounded in the source note and organized around the predicted diagnoses. Focus on the presenting problem, the major active inpatient problems, key treatments or procedures, the clinical course, and the discharge status. Use discharge-summary style prose. Do not copy the note verbatim. Do not invent unsupported diagnoses, procedures, outcomes, or discharge plans. Keep the summary concise, clinically specific, and faithful to the hospitalization.
\end{tcolorbox}
 
\begin{tcolorbox}[promptbox, title=BHC Judge]
You are a senior hospitalist evaluating multiple candidate Brief Hospital Course (BHC) summaries for the same hospitalization.
 
\medskip
\textbf{Source Note} (the original discharge-diagnosis section has been removed)\textbf{:} \texttt{\{clinical\_note\}}
 
\medskip
\textbf{Reference BHC:} \texttt{\{target\_bhc\}}
 
\medskip
\textbf{Anonymous Candidate Systems:} \texttt{\{candidate\_systems\_text\}}
 
\medskip
\textbf{Task:}
Evaluate all candidate BHCs using both the source note and the reference BHC. Rank all systems on the following dimensions: faithfulness to the source note, agreement with the reference BHC, clinical prioritization, timeline coherence, diagnosis-treatment linkage, complication and outcome tracking, information completeness, conciseness and information density, clinical readability, and discharge utility. For each dimension, rank every system with 1 as the best. Then provide an overall ranking from best to worst.
\end{tcolorbox}

\section{Panel Size Sensitivity and Selection Rationale}
\label{sec:panel_size}

\begin{figure}[h]
    \centering
    \includegraphics[width=\linewidth]{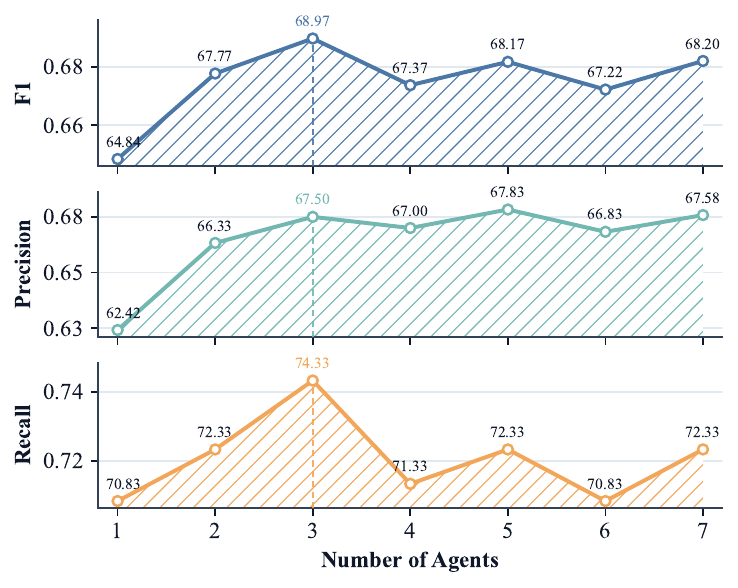}
    \caption{Effect of specialist panel size on diagnostic prediction (evaluated on a 200-case subset). The dashed line marks where performance peaks.}
    \label{fig:agent_num}
\end{figure}

Figure~\ref{fig:agent_num} visualizes the trends reported in Table~\ref{tab:agent_metrics}. The improvement from one to three agents is consistent across all three metrics, with precision showing the largest relative gain (+8.1\%), suggesting that additional specialists primarily help filter false-positive diagnoses. Beyond three agents, all metrics fluctuate within a narrow band (F1 between 0.672 and 0.682 for $k=4-7$), with no statistically meaningful upward trend. We adopt $k=3$ as the default in all other experiments.

\begin{table}[t]
\centering
\small
\caption{Per-note token consumption and diagnostic prediction performance on Llama-3.1-70B.}
\setlength{\tabcolsep}{5pt}
\label{tab:token}
\begin{tabular}{lrrr}
\toprule
\rowcolor{blue}
\camphead{Method} & \camphead{Tokens} & \camphead{Macro F1} & \camphead{Perfect Rate} \\
\midrule
\rowcolor{bluelight}
CAMP (Ours) & 15,458 & \campmetric{0.8750} & \campmetric{67.10} \\
Majority Voting & 6,252 & 0.8584 & 66.55 \\
LLM-as-a-Judge & 17,249 & 0.8313 & 57.40 \\
Devil's Advocate & 18,673 & 0.8306 & 57.20 \\
MedAgents & 19,721 & 0.8277 & 56.70 \\
Chain-of-Thought & \campmetric{1,854} & 0.7899 & 45.55 \\
Self-Consistency & 19,079 & 0.7891 & 45.05 \\
Single Agent & 1,908 & 0.7868 & 44.25 \\
\bottomrule
\vspace{-20pt}
\end{tabular}
\end{table}

\section{Per-Method Token Consumption Breakdown}
\label{sec:cost_supp}

Table~\ref{tab:token} reports the average per-note token consumption alongside diagnostic prediction performance for all methods on Llama-3.1-70B. The comparison reveals that the relationship between token budget and prediction quality is far from linear; how tokens are allocated matters more than how many are consumed. Self-Consistency and MedAgents both use approximately 19,000 tokens per note, yet MedAgents achieves an F1 of 0.8277 compared to Self-Consistency's 0.7891, a gap of nearly four points at equivalent cost. This discrepancy arises because MedAgents distributes tokens across role-differentiated agents that contribute distinct diagnostic perspectives, whereas Self-Consistency draws multiple paths from the same prompt distribution, yielding redundant reasoning traces. A similar contrast emerges between CAMP and Devil's Advocate: CAMP consumes roughly 3,200 fewer tokens (15,458 vs.\ 18,673) while achieving a 4.4-point F1 advantage (0.8750 vs.\ 0.8306). The efficiency gain stems from two sources. Together, these results show that CAMP occupies a favorable position in the cost--performance trade-off, achieving the highest F1 among all methods while maintaining a token footprint below that of most competing multi-agent approaches.

\onecolumn
\section{CAMP Example Outputs}
\label{app:case_studies}
 
Case 10144089-DS-20 and Case 10330554-DS-13 illustrate the pipeline's ability to resolve clinical term ambiguity and diagnostic conflicts through multi-specialist consensus.

\begin{tcolorbox}[casebox, breakable, title=Case 10144089-DS-20: \\ Subarachnoid Hemorrhage with Anatomical Ambiguity]
\label{box:case_sah}
 
\textbf{Case Summary.}
The patient (F) presented with unresponsiveness and agonal breathing after a fall. Transfer notes show SBPs up to 240s. CT demonstrated a large left-sided aneurysm with diffuse hemorrhage; subsequent CTA confirmed an 11\,mm midline shift and signs of herniation (newly blown right pupil). Clinical findings strongly support diagnoses of SAH and cerebral edema with mass effect.
 
\medskip
\textbf{Gold Diagnoses:} \texttt{Subarachnoid hemorrhage}, \texttt{Left MCA aneurysm}, \texttt{Cerebral edema with compression}.
 
\bigskip
 
\begin{tcolorbox}[votebox, title={\normalfont\textbf{Subarachnoid hemorrhage} \hfill \\ \textsc{Consensus: Keep}}]
\textbf{Neurosurgeon} $\rightarrow$ \texttt{KEEP}\,(1.0)\;---\;
``CTA again demonstrates\ldots\ interval increase in hemorrhage in lateral, [third] and [fourth] ventricles.''
\emph{Extensive intraventricular blood is a hallmark of SAH from a large ruptured aneurysm.}
 
\smallskip
\textbf{Neurointensivist} $\rightarrow$ \texttt{KEEP}\,(1.0)\;---\;
``CT scan demonstrated a large (2.5\,$\times$\,3.5) [aneurysm] with [hemorrhage].''
\emph{Acute onset with GCS 3T and imaging findings confirm a catastrophic SAH.}
 
\smallskip
\textbf{Vascular Neurologist} $\rightarrow$ \texttt{KEEP}\,(1.0)\;---\;
``Diffuse subarachnoid hemorrhage along bilateral cerebral hemispheres.''
\emph{The diffuse pattern of hemorrhage is consistent with a high-pressure aneurysmal rupture.}
\end{tcolorbox}
 
\medskip
 
\begin{tcolorbox}[conflictbox, title={\normalfont\textbf{Left MCA aneurysm} \hfill \\ \textsc{Conflict $\rightarrow$ Arbitration: Accept}}]
\textbf{Neurosurgeon} $\rightarrow$ \texttt{KEEP}\,(1.0)\;---\;
``CTA again demonstrates a large L-sided anterior aneurysm.''
\emph{Clinically, a large aneurysm in the anterior circulation is categorized as MCA or ICA terminus; supports this classification.}
 
\smallskip
\textbf{Neurointensivist} $\rightarrow$ \texttt{REMOVE}\,(1.0)\;---\;
``Large (2.5\,$\times$\,3.5) left sided aneurysm.''
\emph{Note only specifies `L-sided anterior'; lacks explicit `MCA' anatomical confirmation, thus removing specific label per strict criteria.}
 
\smallskip
\textbf{Vascular Neurologist} $\rightarrow$ \texttt{KEEP}\,(0.9)\;---\;
``CTA again demonstrates a large L-sided anterior aneurysm.''
\emph{Based on bleeding distribution and location, this `anterior' aneurysm is functionally equivalent to left MCA territory involvement.}
 
\smallskip
\tcblower
\textbf{Attending Arbitration:} The neurosurgeon and vascular neurologist provide converging anatomical reasoning that ``L-sided anterior'' maps to MCA territory given the hemorrhage distribution. The neurointensivist's objection reflects strict labeling criteria but does not provide counter-evidence against MCA localization. $\rightarrow$~\textbf{INCLUDE}.
\end{tcolorbox}
 
\medskip
 
\begin{tcolorbox}[votebox, title={\normalfont\textbf{Cerebral edema with compression} \hfill \\ \textsc{Consensus: Keep}}]
\textbf{Neurosurgeon} $\rightarrow$ \texttt{KEEP}\,(1.0)\;---\;
``Increase in midline shift to approximately 11\,mm.''
\emph{Severe midline shift is direct evidence of mass effect and cerebral edema causing compression.}
 
\smallskip
\textbf{Neurointensivist} $\rightarrow$ \texttt{KEEP}\,(1.0)\;---\;
``Newly blown right pupil with left pupil 2\,mm and fixed.''
\emph{The abnormal pupillary changes (herniation signs) confirm neurologic decline from compressive edema.}
 
\smallskip
\textbf{Vascular Neurologist} $\rightarrow$ \texttt{KEEP}\,(0.9)\;---\;
``Loaded with mannitol and transferred.''
\emph{Clinical use of mannitol protocol corroborates the presence of severe compressive edema.}
\end{tcolorbox}
 
\medskip
 
\begin{tcolorbox}[distractorbox, title={\normalfont\textbf{Vestibular migraine} \hfill \\ \textsc{Consensus: Remove}~{\normalfont(Distractor)}}]
\textbf{Neurosurgeon} $\rightarrow$ \texttt{REMOVE}\,(1.0)\;---\;
\emph{No mention of migraine history; current herniation signs preclude this diagnosis.}
 
\smallskip
\textbf{Neurointensivist} $\rightarrow$ \texttt{REMOVE}\,(1.0)\;---\;
\emph{GCS 3T and intubated status are incompatible with vestibular clinical presentation.}
 
\smallskip
\textbf{Vascular Neurologist} $\rightarrow$ \texttt{REMOVE}\,(1.0)\;---\;
``GCS 3T with absent cough and gag reflexes.''
\emph{Severity of acute neurological collapse is inconsistent with the benign nature of migraine.}
\end{tcolorbox}
 
\bigskip
\textbf{Final Output:} \texttt{\{Cerebral edema with compression, Left MCA aneurysm, Subarachnoid hemorrhage\}}.
\end{tcolorbox}
 
\bigskip
 
\begin{tcolorbox}[casebox, breakable, title=Case 10330554-DS-13: \\ Gait Instability with Vascular and Alcohol Etiologies]
\label{box:case_vascular}
 
\textbf{Case Summary.}
A 65-year-old male with a history of HTN, DM, and chronic ETOH abuse presents with acute unsteadiness. The patient admits to vodka consumption at 12\,AM and 4\,AM. Exam reveals a wide-based gait with stumbling to the right. Imaging shows nonvisualization of the right vertebral artery and focal left ICA stenosis. The diagnostic focus is differentiating between ethanol-induced ataxia and posterior circulation insufficiency.
 
\medskip
\textbf{Gold Diagnoses:} \texttt{Gait difficulty, likely related to alcohol use}, \texttt{Right vertebral artery occlusion}, \texttt{Left ICA stenosis}.
 
\bigskip
 
\begin{tcolorbox}[conflictbox, title={\normalfont\textbf{Gait difficulty, likely related to alcohol use} \hfill \\ \textsc{Conflict $\rightarrow$ Arbitration: Accept}}]
\textbf{Neurologist} $\rightarrow$ \texttt{KEEP}\,(0.7)\;---\;
``Gait: Wide based, stumbles to the right.''
\emph{While the right-sided stumbling is focal, the wide-based gait in the setting of active ETOH intake strongly suggests ethanol-related cerebellar impact.}
 
\smallskip
\textbf{Vascular Surgeon} $\rightarrow$ \texttt{REMOVE}\,(0.6)\;---\;
``Nonvisualization of the right vertebral artery\ldots\ likely due to occlusion.''
\emph{The patient's specific tendency to `stumble to the right' aligns anatomically with the right vertebral artery pathology rather than systemic intoxication.}
 
\smallskip
\textbf{Addiction Specialist} $\rightarrow$ \texttt{KEEP}\,(0.9)\;---\;
``He awoke around 4\,am\ldots\ each time he had a drink or two of vodka.''
\emph{Active consumption less than one hour prior to symptoms directly correlates with the onset of acute ataxia.}
 
\smallskip
\tcblower
\textbf{Attending Arbitration:} The temporal correlation between acute alcohol consumption and symptom onset is the strongest evidence. The vascular surgeon's lateralization argument, while plausible, is lower-confidence (0.6) and speculative. The neurologist and addiction specialist converge on the ethanol etiology. $\rightarrow$~\textbf{INCLUDE}.
\end{tcolorbox}
 
\medskip
 
\begin{tcolorbox}[votebox, title={\normalfont\textbf{Right vertebral artery occlusion} \hfill \\ \textsc{Consensus: Keep}}]
\textbf{Neurologist} $\rightarrow$ \texttt{KEEP}\,(0.8)\;---\;
``CTA HEAD/NECK: Nonvisualization of the right vertebral artery from its origin.''
\emph{Imaging evidence supports occlusion; although MRI showed no acute infarct, chronic occlusion impacts global perfusion.}
 
\smallskip
\textbf{Vascular Surgeon} $\rightarrow$ \texttt{KEEP}\,(1.0)\;---\;
``Minimal retrograde flow into the V4 segment of the right vertebral artery.''
\emph{Nonvisualization at the origin with distal retrograde flow is pathognomonic for total occlusion; requires hemodynamic assessment.}
 
\smallskip
\textbf{Addiction Specialist} $\rightarrow$ \texttt{NEUTRAL}\,(0.0)\;---\;
\emph{Interpretation of complex neurovascular flow patterns is outside the scope of addiction medicine expertise.}
\end{tcolorbox}
 
\medskip
 
\begin{tcolorbox}[votebox, title={\normalfont\textbf{Left ICA stenosis} \hfill \\ \textsc{Consensus: Keep}}]
\textbf{Neurologist} $\rightarrow$ \texttt{KEEP}\,(0.8)\;---\;
``Focal narrowing of the distal ICA just before the entrance into the carotid canal.''
\emph{Significant narrowing identifies a high-risk vascular profile, necessitating secondary stroke prevention.}
 
\smallskip
\textbf{Vascular Surgeon} $\rightarrow$ \texttt{KEEP}\,(0.9)\;---\;
``Focal narrowing\ldots\ consideration should be given to dissection.''
\emph{Critical segment narrowing requires MRI/MRA to rule out acute dissection, which would fundamentally change the surgical management.}
 
\smallskip
\textbf{Addiction Specialist} $\rightarrow$ \texttt{NEUTRAL}\,(0.0)\;---\;
\emph{Large vessel stenosis is a structural vascular diagnosis and does not impact the immediate detoxification or substance abuse treatment plan.}
\end{tcolorbox}
 
\medskip
 
\begin{tcolorbox}[distractorbox, title={\normalfont\textbf{Left frontal SDH} \hfill \\ \textsc{Consensus: Remove}~{\normalfont(Distractor)}}]
\textbf{Neurologist} $\rightarrow$ \texttt{REMOVE}\,(1.0)\;---\;
``CT HEAD: No evidence of acute intracranial process.''
\emph{Acute hemorrhage is radiographically excluded.}
 
\smallskip
\textbf{Vascular Surgeon} $\rightarrow$ \texttt{REMOVE}\,(1.0)\;---\;
\emph{No vascular or traumatic evidence to support a subdural collection.}
 
\smallskip
\textbf{Addiction Specialist} $\rightarrow$ \texttt{REMOVE}\,(1.0)\;---\;
\emph{Clinical presentation is more consistent with ETOH effects than trauma-induced mass effect.}
\end{tcolorbox}
 
\bigskip
\textbf{Final Output:} \texttt{\{Gait difficulty, likely related to alcohol use, Left ICA stenosis, Right vertebral artery occlusion\}}.
\end{tcolorbox}

\end{document}